\documentclass{article}
\usepackage{ijcai22}

\usepackage{times}
\usepackage{soul}
\usepackage{url}
\usepackage[utf8]{inputenc}
\usepackage[small]{caption}
\usepackage{graphicx}
\usepackage{amsmath}
\usepackage{amsthm}
\usepackage{booktabs}
\usepackage{algorithm}
\usepackage{algorithmic}
\usepackage{setspace}
\usepackage{bm}
\usepackage{multirow}
\usepackage{pifont}
\usepackage{balance}
\usepackage{makecell}
\usepackage{array}
\usepackage{booktabs}
\usepackage{amssymb} 
\usepackage{dsfont}  
\usepackage{colortbl} 
\usepackage{cuted} 
\definecolor{mygray}{gray}{0.93}

\title{Beyond the Prototype: Divide-and-conquer Proxies for Few-shot Segmentation}

\author{
Chunbo Lang\footnotemark[1]
\and
Binfei Tu\footnotemark[1]\and
Gong Cheng\footnotemark[2]\and
Junwei Han

\affiliations
{School of Automation, Northwestern Polytechnical University, Xi'an, China}
\emails
\{langchunbo,\,binfeitu\}@mail.nwpu.edu.cn,
\{gcheng,\,jhan\}@nwpu.edu.cn
}

\begin{document}

\maketitle
\renewcommand{\thefootnote}{\fnsymbol{footnote}}
\footnotetext[1]{Equal contribution.}
\footnotetext[2]{Corresponding author.}
\begin{abstract} 
Few-shot segmentation, which aims to segment unseen-class objects given only a handful of densely labeled samples, has received widespread attention from the community. Existing approaches typically follow the prototype learning paradigm to perform meta-inference, which fails to fully exploit the underlying information from support image-mask pairs, resulting in various segmentation failures, \textit{\textit{e.g.}}, incomplete objects, ambiguous boundaries, and distractor activation. To this end, we propose a simple yet versatile framework in the spirit of divide-and-conquer. Specifically, a novel self-reasoning scheme is first implemented on the annotated support image, and then the coarse segmentation mask is \textit{divided} into multiple regions with different properties. Leveraging effective masked average pooling operations, a series of support-induced proxies are thus derived, each playing a specific role in \textit{conquering} the above challenges. Moreover, we devise a unique parallel decoder structure that integrates proxies with similar attributes to boost the discrimination power. Our proposed approach, named divide-and-conquer proxies (DCP), allows for the development of appropriate and reliable information as a guide at the ``episode'' level, not just about the object cues themselves. Extensive experiments on PASCAL-$5^i$ and COCO-$20^i$ demonstrate the superiority of DCP over conventional prototype-based approaches (up to $5\,$$\sim$$10\%$ on average), which also establishes a new state-of-the-art. Code is available at \href{https://github.com/chunbolang/DCP}{\textcolor[RGB]{236,0,140}{\textit{github.com/chunbolang/DCP}}}.
\end{abstract}

\section{Introduction}
Benefiting from the superiority of deep convolutional neural networks, various computer vision tasks have made tremendous progress over the past few years, including image classification \cite{he2016deep}, object detection \cite{ren2015faster}, and semantic segmentation \cite{long2015fully}, to name a few. However, such an effective technique also has its inherent limitation: the strong demand for a considerable number of annotated samples from well-established datasets to achieve satisfactory performance \cite{deng2009imagenet}. For the tasks such as semantic segmentation that performs dense predictions on given images, the acquisition of training sets with sufficient data costs even more, substantially hindering the development of deep learning systems. Few-shot learning (FSL) has emerged as a promising research direction to address this issue, which intends to learn a generic model that can recognize new concepts with very limited information available \cite{vinyals2016matching} \cite{lang2022bam}. 
\begin{figure}[t]
	\centering
	\setlength{\abovecaptionskip}{5pt}
	\includegraphics[width=0.90\linewidth]{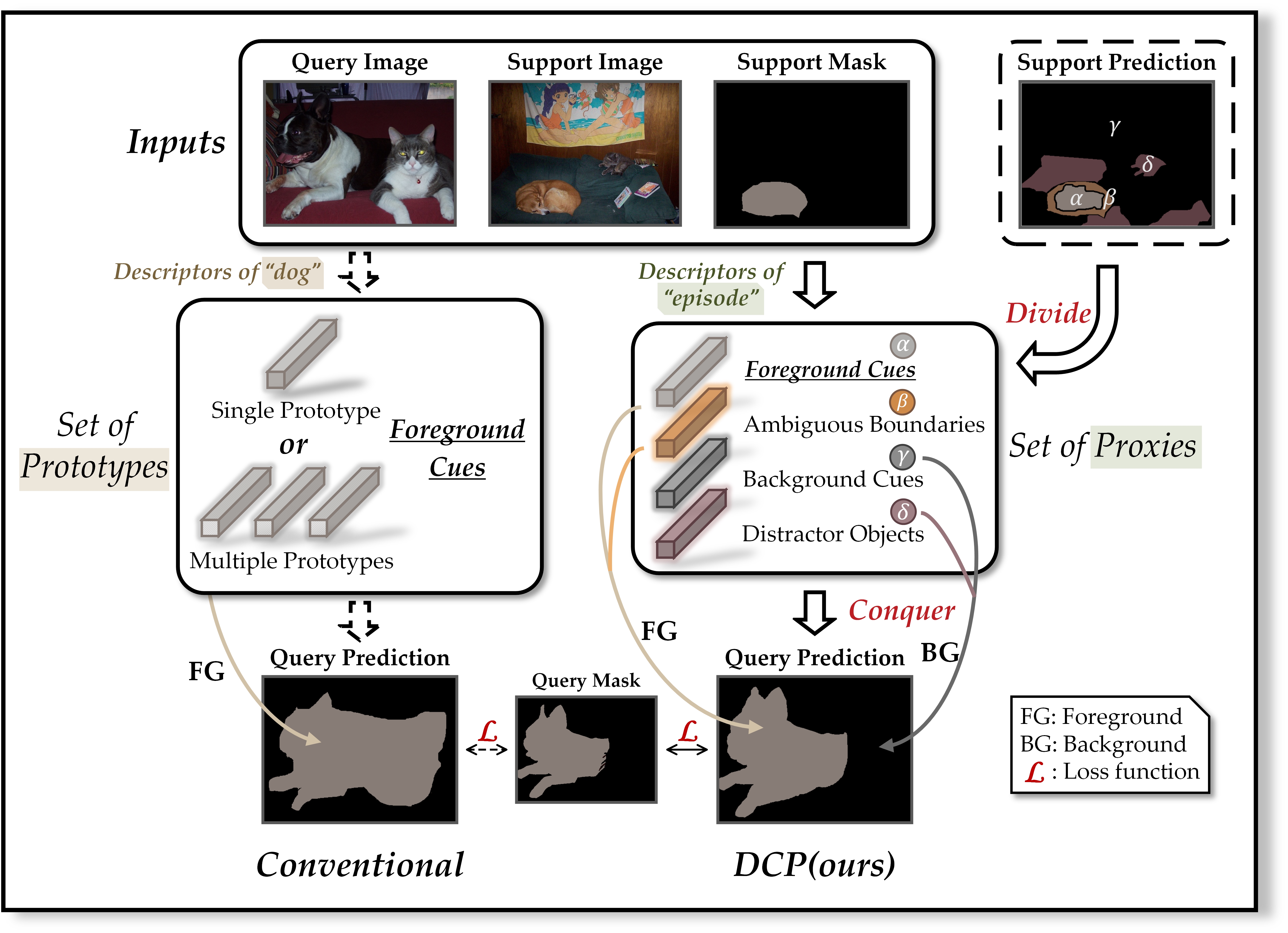}
	\caption{Comparison of conventional FSS approaches (left part) and our DCP approach (right part). \textbf{Conventional approaches} typically leverage support features and corresponding masks to generate single or multiple foreground \textit{prototypes} for guiding query image segmentation, while \textbf{DCP} utilizes the visual reasoning results of support images to derive many different ``\textit{proxies}'' from a broader episodic perspective (divide), each of which plays a specific role (conquer). Overall, such operations could provide pertinent and reliable guidance, including foreground/background cues, ambiguous boundaries, and distractor objects, rather than merely object-related information.}
	\label{fig:1}
\end{figure}
\\\indent In this paper, we undertake the few-shot segmentation (FSS) task that can be regarded as a natural application of FSL technology in the field of image segmentation \cite{shaban2017one}. Specifically, FSS models aim to retrieve the foreground regions of a specific category in the query image while only providing a handful of support images with corresponding annotated masks. Prevalent FSS approaches \cite{zhang2019canet,tian2020prior} follow the prototype learning paradigm to perform meta-reasoning, where single or multiple prototypes are generated by the average pooling operation \cite{zhang2018sg}, furnishing query object inference with essential foreground cues. However, their segmentation capabilities could be fragile when confronted with daunting few-shot tasks with factors such as blurred object boundaries and irrelevant object interference. We argue that one possible reason for these failures lies in the underexplored information from support image-mask pairs, and it is far from enough to rely solely on squeezed object-related features to guide segmentation. Taking the 1-shot segmentation task in Figure\,\textcolor[rgb]{1,0,0}{\ref{fig:1}} as an example, conventional approaches tend to confuse several distractor objects (\textit{e.g.}, \textit{cat} and \textit{sofa}) and yield inaccurate segmentation boundaries (\textit{e.g.}, \textit{dog}’s paw and head), revealing the limitations of the scheme based on such descriptors.\\
\indent To alleviate the abovementioned problems, we propose a simple yet versatile framework in the spirit of divide-and-conquer (refer to the right part of Figure\,\textcolor[rgb]{1,0,0}{\ref{fig:1}}), where the descriptors for guidance are derived from a broader \textit{episodic} perspective, rather than just the objects themselves. Concretely, we first implement a novel self-reasoning scheme on the annotated support image, and then \textit{divide} the coarse segmentation mask into multiple regions with different properties. Using effective masked average pooling operations, a series of support-induced proxies are thus generated, each playing a specific role in \textit{conquering} the typical challenges of daunting FSS tasks. Furthermore, we devise a unique parallel decoder structure in which proxies with similar attributes are integrated to boost the discrimination power. Compared with conventional prototype-based approaches, our proposed framework, named divide-and-conquer proxies (DCP), allows for the development of appropriate and reliable information as a guide at the ``episode'' level, not just about the object cues. \\
\indent Our primary contributions can be summarized as follows: \\
\indent $\bullet\;$ We present a simple yet versatile FSS framework in the spirit of divide-and-conquer, where the descriptors for guidance are derived from a broader episodic perspective to tackle the typical challenges of daunting tasks.\\ 
\indent $\bullet\;$Compared with conventional prototype-based approaches, the proposed framework allows for the development of appropriate and reliable information, rather than merely the object cues themselves, providing a fresh insight for future works.\\
\indent $\bullet\;$ Extensive experiments on two standard benchmarks demonstrate the superiority of our framework over the baseline approach (up to 5$\sim$10\% on average), establishing new state-of-the-arts in few-shot literature. 
\begin{figure*}[t]
	\centering
	\includegraphics[width=0.9\linewidth]{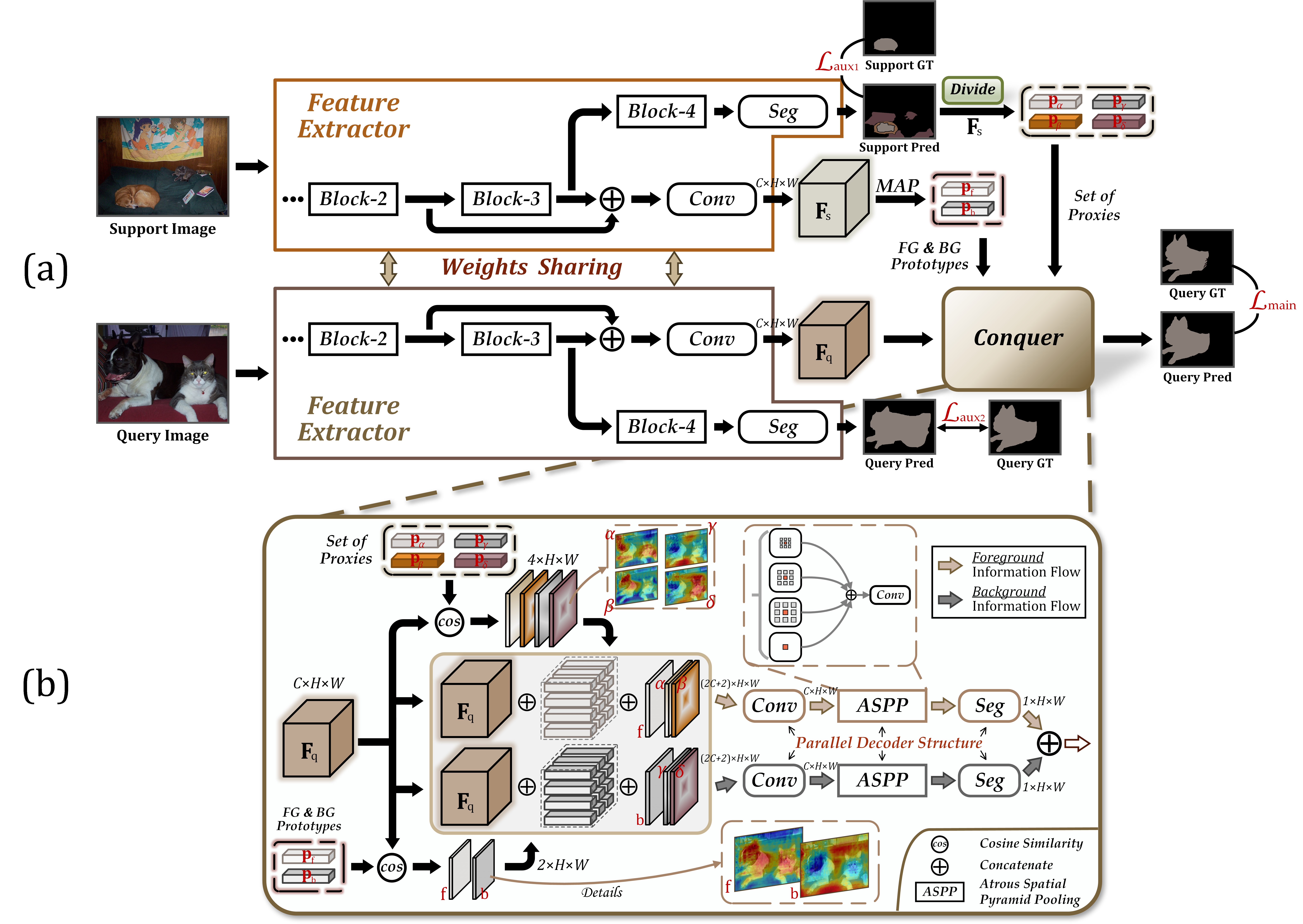}
	\caption{DCP for 1-shot semantic segmentation. (a) Overall pipeline of the proposed approach. (b) Implementation details of the conquer module.}
	\label{fig:2}
\end{figure*}
\section{Related Works}
\label{sec:2}
\textbf{Semantic Segmentation.} Semantic segmentation is a fundamental and essential computer vision task, the goal of which is to classify each pixel in the image according to a predefined set of semantic categories. Most advanced approaches follow fully convolutional networks (FCNs) \cite{long2015fully} to perform dense predictions, vastly boosting the segmentation performance \cite{zhao2017pyramid}. On top of that, the rapid development of this field has also brought some impressive techniques, such as dilated convolutions \cite{chen2017deeplab} and attention mechanism \cite{huang2019ccnet}. In this work, we adopt the Atrous Spatial Pyramid Pooling (ASPP) module \cite{chen2017deeplab} based on dilated convolution to build the decoder, enjoying a large receptive field and multi-scale feature aggregation. \\
\textbf{Few-shot Learning.} Few-shot learning (FSL) is proposed to tackle the generalization problem of conventional deep networks, where meta-knowledge is shared across tasks to facilitate the recognition of new concepts with very little support information. Representative FSL approaches can be grouped into three categories: optimization-based method \cite{finn2017model}, hallucination-based method \cite{chen2019imaged}, and metric-based method \cite{vinyals2016matching}. Our work closely relates to the prototypical network (PN) \cite{snell2017prototypical} assigned to the third category. PN directly leverages the feature representations (\textit{i.e.}, prototypes) computed through global average pooling operations to perform \textit{non-parametric} nearest neighbor classification. Our DCP, on the other hand, utilizes foreground/background prototypes for dense feature comparison \cite{zhang2019canet}, providing essential information for subsequent \textit{parametric} query image segmentation. \\
\textbf{Few-shot Segmentation.} Few-shot segmentation (FSS) has emerged as a promising research direction to tackle the dense prediction problem in a low-data regime. Existing approaches typically employ a two-branch structure, in which the annotation information travels from the support branch to the query branch \cite{zhang2019canet} \cite{siam2019amp}. Generally speaking, how to achieve more effective information interaction between the two branches is the primary concern in this field. OSLSM \cite{shaban2017one}, the pioneering work of FSS, proposed to generate classifier weights for query image segmentation in the conditioning (support) branch. Whereafter, the prototype learning paradigm \cite{snell2017prototypical} was introduced into state-of-the-art approaches for better information exchange. SG-One \cite{zhang2018sg} computed the affinity scores between the prototype and query features, yielding the spatial similarity map as a guide. CANet \cite{zhang2019canet} proposed to conduct a novel dense comparison operation based on the prototype, which exhibited decent segmentation performance under few-shot settings. However, these methods are limited to the utilization of features of foreground objects (\textit{i.e.}, prototype) to facilitate segmentation and fail to provide pertinent and reliable guidance from a broader \textit{episodic} perspective. Our method draws on the idea of prototype learning while proposing a set of support-induced proxies that go beyond it. 

\section{Problem Setup}
\label{sec:3}
Few-shot segmentation (FSS) aims to segment the objects of a specific category in the query image using only a few labeled samples. To improve the generalization ability for unseen classes, existing approaches widely adopt the meta-learning strategy for training, also known as episodic learning \cite{vinyals2016matching}. Specifically, given a training set $\mathcal{D_{{\rm{train}}}}$ and a testing set $\mathcal{D_{{\rm{test}}}}$ that are disjoint with regard to object classes, a series of \textit{episodes} are sampled from them to simulate the few-shot scenario. For a $K$-shot segmentation task, each episode is composed of \textbf{1)} a support set $\mathcal{S} = \{{({{\bf{I}}_{\rm{s}}^k,{\bf{M}}_{{\rm{s}},c}^k})}\}_{c \in {\mathcal{C}_{{\rm{episode}}}}}^{k = 1,...,K}$, where ${\bf{I}}_{\rm{s}}^k$ is the $k$-th support image, ${\bf{M}}_{{\rm{s}},c}^k$ is the corresponding mask for class $c$, and ${\mathcal{C}_{{\rm{episode}}}}$ represents the class set of the given episode; and \textbf{2)} a query set $\mathcal{Q} = \{{{{\bf{I}}_{\rm{q}}},{{\bf{M}}_{{\rm{q}},c}}}\}$ where ${{\bf{I}}_{\rm{q}}}$ is the query image and ${{\bf{M}}_{{\rm{q}},c}}$ is the ground-truth mask \textit{available} during training while \textit{unknown} during testing. 
\section{Method}
\label{sec:4}
In this section, we propose a simple yet effective few-shot segmentation framework, termed divide-and-conquer proxies (DCP). The unique feature of our approach, as also the source of improvements, lies in the spirit of divide-and-conquer. The \textit{divide} module generates a set of proxies according to the prediction and ground-truth mask of the support image; the \textit{conquer} module utilizes these distinguishing proxies to provide essential segmentation cues for the query branch, as depicted in Figure\,\textcolor[rgb]{1,0,0}{\ref{fig:2}}. 
\subsection{Divide}
\begin{figure}[t]
	\hspace{-0.5cm}
	\setlength{\abovecaptionskip}{4pt}
	\centering
	\includegraphics[width=0.95\linewidth]{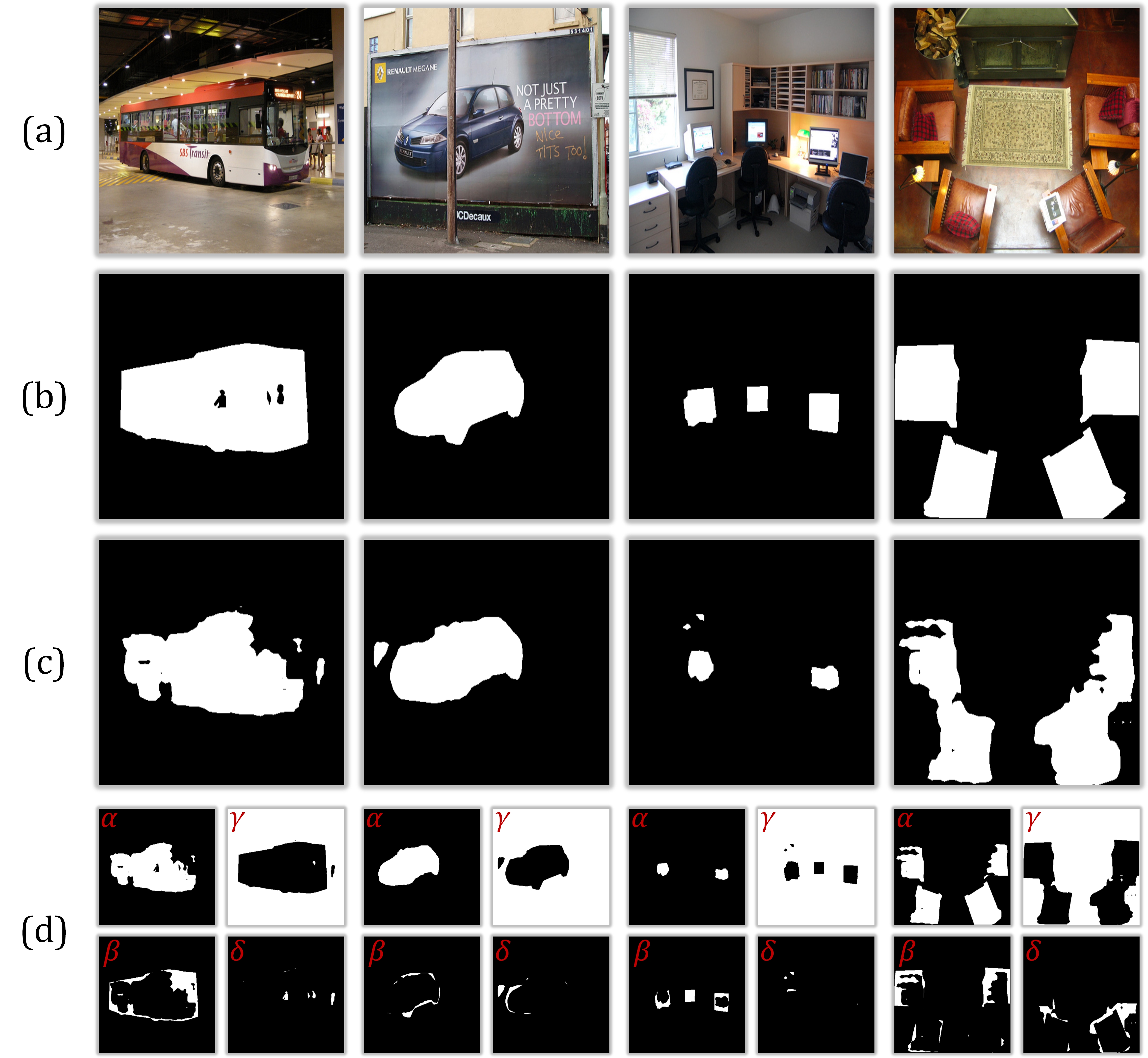}
	\caption{Visualization of the masks for generating proxies. From top to bottom, each row represents (a) raw (support) images, (b) ground-truth masks, (c) prediction masks, and (d) masks for generating various proxies, respectively.}
	\label{fig:3}
\end{figure}

\setlength {\parskip} {0pt}
\label{sec:4.1}
\textbf{Self-reasoning Scheme.} Following the paradigm of advanced FSS approaches \cite{zhang2019canet}, we freeze the backbone network to boost generalization capabilities. Taking ResNet \cite{he2016deep} for example, the intermediate feature maps after each convolutional block of the support branch are represented as $\{{{\bf{F}}_{\rm{s}}^b}\}_{b = 1}^B$\textcolor[rgb]{1,0,0}{\footnote{$B$ denotes the total number of convolutional blocks contained in the backbone network, which is equal to 4 for ResNet.}}. We use the high-level features generated by the last block, \textit{i.e.}, \textit{block}4, to perform the self-reasoning scheme:
\begin{equation}
{\bf{\hat M}}_{\rm{s}}^{{\rm{aux}}}{\rm{ = }}{f_{{\rm{seg}}}}( {{\bf{F}}_{\rm{s}}^4} ),
\end{equation}
where ${f_{{\rm{seg}}}}(\cdot)$ is a lightweight segmentation decoder with residual connections, including three convolutions with 256 filters, one convolution with 2 filters, and an $\arg \max (\cdot)$ operation for producing coarse prediction masks ${\bf{\hat M}}_{\rm{s}}^{{\rm{aux}}}$$\in$${\{{0,1}\}^{H \times W}}$. $H$ and $W$ denote the height and weight respectively, and their product $H$$\times$$W$ is the minimum resolution of all feature maps. Unlike the widely adopted prototype-guided segmentation framework, such a \textit{mask-agnostic} self-reasoning scheme is more efficient and can also provide reliable auxiliary information.
\\
\textbf{Proxy Acquisition.} Given the coarse prediction ${\bf{\hat M}}_{\rm{s}}^{{\rm{aux}}}$ and down-sampled ground-truth mask ${{\bf{M}}_{\rm{s}}}$ of the support image, we first evaluate their differences to derive the corresponding mask (\textit{i.e.}, valid region) ${{\bf{M}}_{m \in {\mathcal{P}_1}}}$ of each proxy where ${\mathcal{P}_1}$$=$$\{{\alpha ,\beta ,\gamma ,\delta }\}$ denotes the index set of proxies: 
\begin{equation}
\left\{ \begin{array}{l}
{\bf{M}}_\alpha ^{\left( {x,y} \right)} = \mathds{1}[ {{\bf{\hat M}}_{\rm{s}}^{{\rm{aux;}}\left( {x,y} \right)} = {\bf{M}}_{\rm{s}}^{\left( {x,y} \right)} = 1}]\\
{{\bf{M}}_\beta } {\kern 2.8ex} = {{\bf{M}}_{\rm{s}}}{\rm{ - }}{{\bf{M}}_\alpha }\\
{\bf{M}}_\gamma ^{\left( {x,y} \right)} = \mathds{1}[ {{\bf{\hat M}}_{\rm{s}}^{{\rm{aux;}}\left( {x,y} \right)} = {\bf{M}}_{\rm{s}}^{\left( {x,y} \right)} = 0}]\\
{{\bf{M}}_\delta } {\kern 3.0ex} = {\bf{G}}{\rm{ - }}{{\bf{M}}_{\rm{s}}}{\rm{ - }}{{\bf{M}}_\gamma }
\end{array} \right.,
\end{equation}
where $({x,y})$ indexes the spatial location of the mask, $\mathds{1}(\cdot)$ is an indicator function that outputs $1$ if the condition is satisfied or $0$ otherwise, and ${\bf{G}}$$\in$$\mathbb{R}{^{H \times W}}$ represents an all $1$'s matrix with the same shape as ${{\bf{M}}_m}$. ${{\bf{M}}_\alpha }$ and ${{\bf{M}}_\beta }$ aggregate the main and auxiliary foreground information respectively, generally corresponding to the body and boundary regions of objects; while ${{\bf{M}}_\gamma }$ and ${{\bf{M}}_\delta }$ gather the primary and secondary background information respectively, which typically correspond to the regions of the generalized background and distractor objects. As described above, these masks are generated according to the segmentation results of the model, therefore the main (primary) and auxiliary (secondary) status may change when encountering complex tasks. The extreme case is ${{\bf{M}}_\beta }$$=$${{\bf{M}}_{\rm{s}}}$, but even so, the guidance information is still complete. To better understand the effect of each mask, we present several examples in Figure\,\textcolor[rgb]{1,0,0}{\ref{fig:3}}. \\
\indent Then, we conduct masked average pooling operations \cite{zhang2018sg} on the support features ${{\bf{F}}_{\rm{s}}}$$\in$$\mathbb{R}{^{C \times H \times W}}$ and masks ${{\bf{M}}_m}$ computed above to encode the set of proxies $\{{{{\bf{p}}_m}}\}$: 
\begin{equation}
\label{eq:3}
{{\bf{p}}_m} = \frac{{\sum\nolimits_{x,y} {{\bf{F}}_{\rm{s}}^{(x,y)} \cdot } {\bf{M}}_m^{(x,y)}}}{{\sum\nolimits_{x,y} {{\bf{M}}_m^{(x,y)}} }},
\end{equation}
where ${{\bf{p}}_m}$$\in$$\mathbb{R}{^C}$ represents a specific proxy in the set, playing some roles in conquering the typical challenges of FSS tasks.\vspace{-10pt}
\vspace{2pt}\\
\textbf{Prototype Acquisition.} Similar to the acquisition method of proxies (Eq.\,(\textcolor[rgb]{1,0,0}{\ref{eq:3}})), the set of prototypes ${\{{{{\bf{p}}_n}}\}_{n \in {\mathcal{P}_2}}}$ are also generated using the masked average pooling operation where ${\mathcal{P}_2}$$=$$\left\{ {{\rm{f}},{\rm{b}}} \right\}$ represents the index set of prototypes:
\begin{equation}
\label{eq:4}
\setlength{\abovedisplayskip}{4pt}
{{\bf{p}}_n} = \frac{{\sum\nolimits_{x,y} {{\bf{F}}_{\rm{s}}^{(x,y)} \cdot \mathds{1}[ {{\bf{M}}_{\rm{s}}^{(x,y)}{\rm{ = }}j} ]} }}{{\sum\nolimits_{x,y} {\mathds{1}[ {{\bf{M}}_{\rm{s}}^{(x,y)}{\rm{ = }}j} ]} }},
\setlength{\belowdisplayskip}{4pt}
\end{equation}
where $j$$=$$1$ if the foreground prototype ${{\bf{p}}_{\rm{f}}}$ is evaluated and $j$$=$$0$ if the background one ${{\bf{p}}_{\rm{b}}}$ is calculated. 

\subsection{Conquer}
\label{sec:4.2}
\begin{figure}[t]
	\vspace{-0.2cm}
	\centering
	\setlength{\abovecaptionskip}{4pt}
	\includegraphics[width=0.95\linewidth]{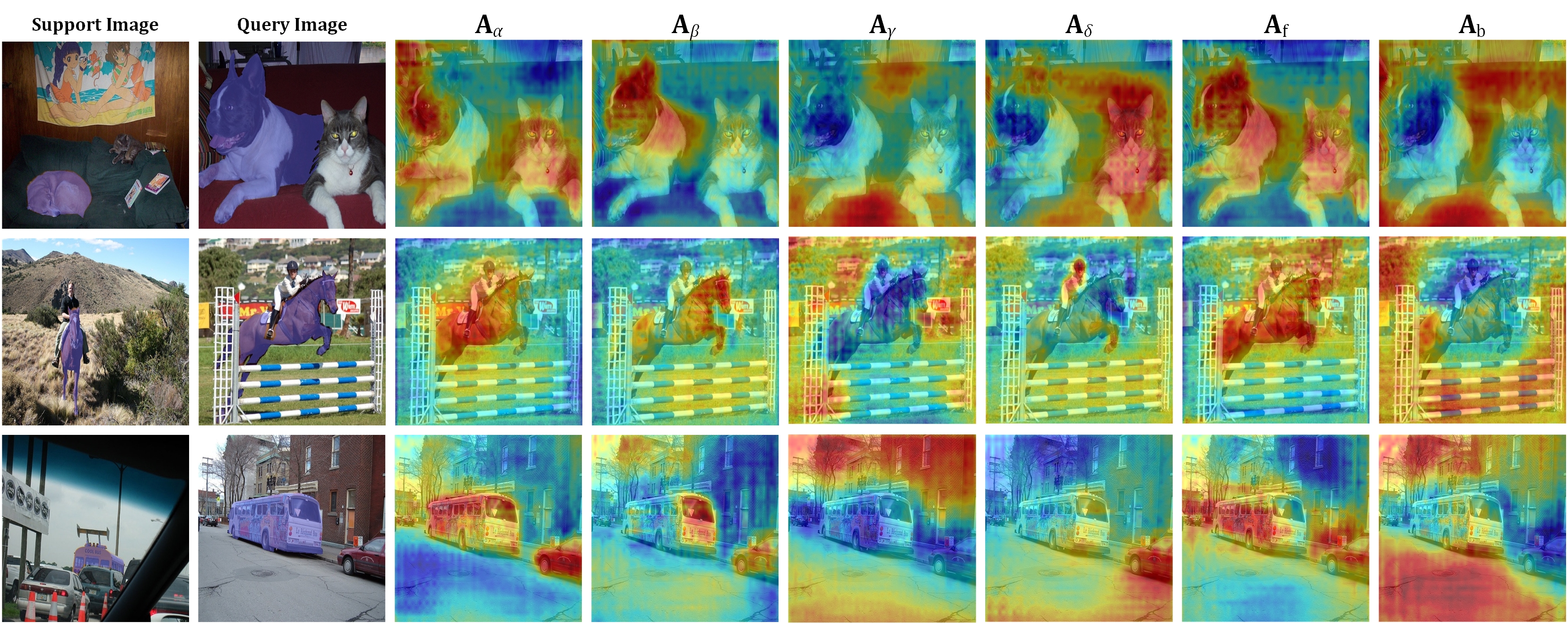}
	\caption{Visualization of activation maps generated by proxies and prototypes. Best viewed in color.}
	\label{fig:4}
	\vspace{0.3cm}
\end{figure}

\textbf{Feature Matching.} Dense feature comparison is a common approach in the FSS literature that compares the query feature maps ${{\bf{F}}_{\rm{q}}}$ with the global feature vector (\textit{i.e.}, prototype) at all spatial locations. In view of its effectiveness, we follow the same technical route and extend it to a dual-prototype setting (see Figure\,\textcolor[rgb]{1,0,0}{\ref{fig:3}}(b)), which can be defined as:
\begin{equation}
\label{eq:5}
{\bf{F}}_{{\rm{q;}}n}^{{\rm{tmp}}}{\rm{ = }} \bigoplus \left( {{{\bf{F}}_{\rm{q}}},{\zeta_{H \times W}}( {{{\bf{p}}_n}} )} \right),
\end{equation}
where ${\zeta_{h \times w}}(\cdot)$ expands the given vector to a spatial size $h$$\times$$w$ and $\bigoplus$ represents the concatenation operation along channel dimensions. The superscript ``tmp'' denotes \textit{temporary}, and the subscript $n$$\in$${\mathcal{P}_2}$ indicates the index of the prototype.
\vspace{2pt}\\
\textbf{Feature Activation.} Given the proxies ${\{{{{\bf{p}}_m}}\}_{m \in {\mathcal{P}_1}}}$ and prototypes ${\{{{{\bf{p}}_n}}\}_{n \in {\mathcal{P}_2}}}$ computed by Eqs.\,(\textcolor[rgb]{1,0,0}{\ref{eq:3}}-\textcolor[rgb]{1,0,0}{\ref{eq:4}}), we evaluate the cosine similarity between each of them and query features at each spatial location respectively: 
\begin{equation}
\label{eq:6}
\setlength{\abovedisplayskip}{4pt}
{\bf{A}}_l^{\left( {x,y} \right)} = \frac{{{\bf{p}}_l^\mathsf{T} \cdot {\bf{F}}_{\rm{q}}^{\left( {x,y} \right)}}}{{\left\| {{{\bf{p}}_l}} \right\| \cdot \left\| {{\bf{F}}_{\rm{q}}^{\left( {x,y} \right)}} \right\|}},\;\;\;l \in {\mathcal{P}_1} \cup {\mathcal{P}_2},
\setlength{\belowdisplayskip}{4pt}
\end{equation}
where ${{\bf{A}}_l}$$\in$$\mathbb{R}^{H \times W}$ denotes the activation map of the $l$-th item. Figure\,\textcolor[rgb]{1,0,0}{\ref{fig:4}} presents some examples to illustrate the region of interest for each feature vector ${{\bf{p}}_l}$. Actually, there is an alternative approach to propagate information about feature vectors ${\{{{{\bf{p}}_l}}\}_{l \in {\mathcal{P}_1} \cup {\mathcal{P}_2}}}$ to the query branch, that is, to perform the same operation as Eq.\,(\textcolor[rgb]{1,0,0}{\ref{eq:5}}). Such a scheme, however, introduces additional computational overhead and performance degradation, as discussed in \S\textcolor[rgb]{1,0,0}{\,\ref{sec:5.3}}.
\vspace{2pt}\\
\textbf{Parallel Decoder Structure.} Most previous works abandon the prototype ${{\bf{p}}_{\rm{b}}}$ due to the complexity of background regions in support images. We argue that specifically designing the network to capture foreground and background features separately could have a positive effect on performance, whereas a simple fusion strategy would be counterproductive. To this end, a unique parallel decoder structure (PDS) is devised in which proxies/prototypes with similar attributes are integrated to boost the discrimination power:
\begin{equation}
\label{eq:7}
\setlength{\abovedisplayskip}{4pt}
{\bf{F}}_{{\rm{q;}}n}^{{\rm{refined}}}{\rm{ = }} \bigoplus \left( {{\bf{F}}_{{\rm{q;}}n}^{{\rm{tmp}}},{{\bf{A}}_n},{{\left\{ {{{\bf{A}}_m}} \right\}}_{m \in {\mathcal{P}_{{\rm{tmp}}}}}}} \right),
\setlength{\belowdisplayskip}{4pt}
\end{equation}
where ${\mathcal{P}_{{\rm{tmp}}}}$ is a temporary set of vector indexes, representing $\{{\alpha ,\beta }\}$ when foreground features ${\bf{F}}_{{\rm{q;f}}}^{{\rm{refined}}}$ are evaluated and $\{{\gamma ,\delta }\}$ when background ones ${\bf{F}}_{{\rm{q;b}}}^{{\rm{refined}}}$ are computed. These two refined query features are then fed into the corresponding decoder network respectively, and the output single-channel probability maps are concatenated to generate the final segmentation result. 

\begin{table*}[t]
	\centering
	\resizebox{0.80\linewidth}{!}{
		\renewcommand{\arraystretch}{1.1}
		\begin{tabular}{|m{1.4cm}<{\centering}|m{5.2cm}<{\raggedright }|m{0.8cm}<{\centering}m{0.8cm}<{\centering}m{0.8cm}<{\centering}m{0.8cm}<{\centering}m{0.8cm}<{\centering}|m{0.8cm}<{\centering}m{0.8cm}<{\centering}m{0.8cm}<{\centering}m{0.8cm}<{\centering}m{0.8cm}<{\centering}|}
			\hline
			\multirow{2}{*}[-0.2ex]{\textbf{Backbone}}  & \multirow{2}{*}[-0.2ex]{\textbf{Method}} & \multicolumn{5}{c|}{\multirow{1}{*}[-0.2ex]{\textbf{1-shot}}} & \multicolumn{5}{c|}{\multirow{1}{*}[-0.2ex]{\textbf{5-shot}}} \\ \cline{3-12} & & \multirow{1}{*}[-0.3ex]{P.-5$^0$} & \multirow{1}{*}[-0.3ex]{P.-5$^1$} & \multirow{1}{*}[-0.3ex]{P.-5$^2$} & \multirow{1}{*}[-0.3ex]{P.-5$^3$} & \multirow{1}{*}[-0.25ex]{Mean}  & \multirow{1}{*}[-0.3ex]{P.-5$^0$} & \multirow{1}{*}[-0.3ex]{P.-5$^1$} & \multirow{1}{*}[-0.3ex]{P.-5$^2$} & \multirow{1}{*}[-0.3ex]{P.-5$^3$} & \multirow{1}{*}[-0.25ex]{Mean}  \\ \hline \hline
			\multirow{6}{*}{\multirow{1}{*}[-0.4ex]{VGG16}}    & OSLSM \cite{shaban2017one} & 33.60 & 55.30 & 40.90 & 33.50 & 40.80 & 35.90 & 58.10 & 42.70 & 39.10 & 43.90 \\
			& FWB \cite{nguyen2019feature} & 47.04 & 59.64 & 52.61 & 48.27 & 51.90 & 50.87 & 62.86 & 56.48 & 50.09 & 55.08 \\
			& PANet \cite{wang2019panet} & 42.30 & 58.00 & 51.10 & 41.20 & 48.10 & 51.80 & 64.60 & \textcolor[rgb]{0,0,1}{\textbf{59.80}} & 46.50 & 55.70 \\
			& PFENet \cite{tian2020prior} & 56.90 & \textcolor[rgb]{0,0,1}{\textbf{68.20}} & 54.40 & \textcolor[rgb]{0,0,1}{\textbf{52.40}} & 58.00 & \textcolor[rgb]{0,0,1}{\textbf{59.00}} & \textcolor[rgb]{0,0,1}{\textbf{69.10}} & 54.80 & 52.90 & \textcolor[rgb]{0,0,1}{\textbf{59.00}} \\
			& MMNet \cite{wu2021learning} & \textcolor[rgb]{0,0,1}{\textbf{57.10}} & 67.20 & \textcolor[rgb]{0,0,1}{\textbf{56.60}} & 52.30 & \textcolor[rgb]{0,0,1}{\textbf{58.30}} & 56.60 & 66.70 & 53.60 & \textcolor[rgb]{0,0,1}{\textbf{56.50}} & 58.30 \\
			& \cellcolor{mygray}DCP (ours) & \cellcolor{mygray}\textcolor[rgb]{1,0,0}{\textbf{59.67}} & \cellcolor{mygray}\textcolor[rgb]{1,0,0}{\textbf{68.67}} & \cellcolor{mygray}\textcolor[rgb]{1,0,0}{\textbf{63.78}} & \cellcolor{mygray}\textcolor[rgb]{1,0,0}{\textbf{53.11}} & \cellcolor{mygray}\textcolor[rgb]{1,0,0}{\textbf{61.31}} & \cellcolor{mygray}\textcolor[rgb]{1,0,0}{\textbf{64.25}} & \cellcolor{mygray}\textcolor[rgb]{1,0,0}{\textbf{70.66}} & \cellcolor{mygray}\textcolor[rgb]{1,0,0}{\textbf{67.38}} & \cellcolor{mygray}\textcolor[rgb]{1,0,0}{\textbf{61.08}} & \cellcolor{mygray}\textcolor[rgb]{1,0,0}{\textbf{65.84}} \\ \hline
			\multirow{6}{*}{\multirow{1}{*}[-0.4ex]{ResNet50}} & CANet \cite{zhang2019canet} & 52.50 & 65.90 & 51.30 & 51.90 & 55.40 & 55.50 & 67.80 & 51.90 & 53.20 & 57.10 \\				
			& SCL$^{\dagger}$ \cite{zhang2021self} & 63.00 & 70.00 & 56.50 & \textcolor[rgb]{1,0,0}{\textbf{57.70}} & 61.80 & 64.50 & 70.90 & 57.30 & 58.70 & 62.90 \\
			& SAGNN \cite{xie2021scale} & \textcolor[rgb]{1,0,0}{\textbf{64.70}} & 69.60 & 57.00 & \textcolor[rgb]{0,0,1}{\textbf{57.20}} & \textcolor[rgb]{0,0,1}{\textbf{62.10}} & \textcolor[rgb]{0,0,1}{\textbf{64.90}} & 70.00 & 57.00 & 59.30 & 62.80 \\
			& MMNet \cite{wu2021learning} & 62.70 & \textcolor[rgb]{0,0,1}{\textbf{70.20}} & 57.30 & 57.00 & 61.80 & 62.20 & \textcolor[rgb]{0,0,1}{\textbf{71.50}} & 57.50 & \textcolor[rgb]{0,0,1}{\textbf{62.40}} & 63.40 \\
			& CWT \cite{lu2021simpler} & 56.30 & 62.00 & \textcolor[rgb]{0,0,1}{\textbf{59.90}} & 47.20 & 56.40 & 61.30 & 68.50 & \textcolor[rgb]{1,0,0}{\textbf{68.50}} & 56.60 & \textcolor[rgb]{0,0,1}{\textbf{63.70}} \\
			& \cellcolor{mygray}DCP (ours) & \cellcolor{mygray}\textcolor[rgb]{0,0,1}{\textbf{63.81}} & \cellcolor{mygray}\textcolor[rgb]{1,0,0}{\textbf{70.54}} & \cellcolor{mygray}\textcolor[rgb]{1,0,0}{\textbf{61.16}} & \cellcolor{mygray}55.69 & \cellcolor{mygray}\textcolor[rgb]{1,0,0}{\textbf{62.80}} & \cellcolor{mygray}\textcolor[rgb]{1,0,0}{\textbf{67.19}} & \cellcolor{mygray}\textcolor[rgb]{1,0,0}{\textbf{73.15}} & \cellcolor{mygray}\textcolor[rgb]{0,0,1}{\textbf{66.39}} & \cellcolor{mygray}\textcolor[rgb]{1,0,0}{\textbf{64.48}} & \cellcolor{mygray}\textcolor[rgb]{1,0,0}{\textbf{67.80}} \\ \hline
	\end{tabular}}
	\caption{Comparison with state-of-the-arts on PASCAL-$5^i$ in mIoU under 1-shot and 5-shot settings. ``P.'' means PASCAL. \textcolor[rgb]{1,0,0}{\textbf{RED}}/\textcolor[rgb]{0,0,1}{\textbf{BLUE}} represents the $1^{\rm st}$/$2^{\rm nd}$ best performance. Superscript ``$\dagger$'' indicates that PFENet is served as the baseline.}
	\label{tab:1}
\end{table*}

\begin{table*}[ht]
	\vspace{0.3cm}
	\begin{minipage}[t]{0.7\textwidth}
		\centering
		\resizebox{0.98\linewidth}{1.05cm}{
			\renewcommand{\arraystretch}{1.05}
			\begin{tabular}{|l|ccccc|ccccc|c|}
				\hline
				\multirow{2}{*}[-0.2ex]{\textbf{Method}} & \multicolumn{5}{c|}{\multirow{1}{*}[-0.2ex]{\textbf{1-shot}}} & \multicolumn{5}{c|}{\multirow{1}{*}[-0.2ex]{\textbf{5-shot}}} &  \multirow{2}{*}[-0.2ex]{\textbf{\#Params.}} \\ \cline{2-11} & \multirow{1}{*}[-0.3ex]{C.-20$^0$} & \multirow{1}{*}[-0.3ex]{C.-20$^1$} & \multirow{1}{*}[-0.3ex]{C.-20$^2$} & \multirow{1}{*}[-0.3ex]{C.-20$^3$} & \multirow{1}{*}[-0.25ex]{Mean}  & \multirow{1}{*}[-0.3ex]{C.-20$^0$} & \multirow{1}{*}[-0.3ex]{C.-20$^1$} & \multirow{1}{*}[-0.3ex]{C.-20$^2$} & \multirow{1}{*}[-0.3ex]{C.-20$^3$} & \multirow{1}{*}[-0.25ex]{Mean}  & \\ \hline \hline
				FWB$^\ddagger$ \cite{nguyen2019feature} & 16.98 & 17.98 & 20.96 & 28.85 & 21.19 & 19.13 & 21.46 & 23.93 & 30.08 & 23.65 & 43.0M \\						
				PFENet$^\ddagger$ \cite{tian2020prior} & \textcolor[rgb]{0,0,1}{\textbf{36.80}} & \textcolor[rgb]{0,0,1}{\textbf{41.80}} & \textcolor[rgb]{0,0,1}{\textbf{38.70}} & 36.70 & \textcolor[rgb]{0,0,1}{\textbf{38.50}} & \textcolor[rgb]{0,0,1}{\textbf{40.40}} & \textcolor[rgb]{0,0,1}{\textbf{46.80}} & \textcolor[rgb]{0,0,1}{\textbf{43.20}} & 40.50 & \textcolor[rgb]{0,0,1}{\textbf{42.70}} & \textcolor[rgb]{0,0,1}{\textbf{10.8M}} \\
				MMNet \cite{wu2021learning} & 34.90 & 41.00 & 37.20 & \textcolor[rgb]{0,0,1}{\textbf{37.00}} & 37.50 & 37.00 & 40.30 & 39.30 & 36.00 & 38.20 & \textcolor[rgb]{1,0,0}{\textbf{10.5M}} \\
				CWT \cite{lu2021simpler} & 32.20 & 36.00 & 31.60 & 31.60 & 32.90 & 40.10 & 43.80 & 39.00 & \textcolor[rgb]{0,0,1}{\textbf{42.40}} & 41.30 & 47.3M \\
				\cellcolor{mygray}DCP (ours) & \cellcolor{mygray}\textcolor[rgb]{1,0,0}{\textbf{40.89}} & \cellcolor{mygray}\textcolor[rgb]{1,0,0}{\textbf{43.77}} & \cellcolor{mygray}\textcolor[rgb]{1,0,0}{\textbf{42.60}} & \cellcolor{mygray}\textcolor[rgb]{1,0,0}{\textbf{38.29}} & \cellcolor{mygray}\textcolor[rgb]{1,0,0}{\textbf{41.39}} & \cellcolor{mygray}\textcolor[rgb]{1,0,0}{\textbf{45.82}} & \cellcolor{mygray}\textcolor[rgb]{1,0,0}{\textbf{49.66}} & \cellcolor{mygray}\textcolor[rgb]{1,0,0}{\textbf{43.69}} & \cellcolor{mygray}\textcolor[rgb]{1,0,0}{\textbf{46.62}} & \cellcolor{mygray}\textcolor[rgb]{1,0,0}{\textbf{46.48}} & \cellcolor{mygray}11.3M \\ \hline
		\end{tabular}}
		\caption{Comparison with state-of-the-arts on COCO-$20^i$ in mIoU under 1-shot and 5-shot settings. ``C.'' means COCO. The methods with superscript ``$\ddagger$'' use the ResNet101 backbone while other approaches use the ResNet50. ``\#Params.'' indicates the number of \textit{learnable} parameters.}
		\label{tab:2}
	\end{minipage}
	\hspace{0.4cm}
	\begin{minipage}[t]{0.25\textwidth}
		\centering
		\resizebox{0.98\linewidth}{1.05cm}{
			\renewcommand{\arraystretch}{1.05}
			\begin{tabular}{|c|l|cc|}
				\hline
				\multirow{2}{*}[-0.2ex]{\textbf{Backbone}} & \multirow{2}{*}[-0.2ex]{\textbf{Method}} & \multicolumn{2}{c|}{\multirow{1}{*}[-0.2ex]{\textbf{FB-IoU}}} \\ \cline{3-4} 
				&  & \multirow{1}{*}[-0.2ex]{1-shot} & \multirow{1}{*}[-0.2ex]{5-shot} \\ \hline \hline
				\multirow{3}{*}{\multirow{1}{*}[-0.4ex]{VGG16}} & OSLSM \cite{shaban2017one} & 61.3 & 61.5 \\
				& PANet \cite{wang2019panet} & \textcolor[rgb]{0,0,1}{\textbf{66.5}} & \textcolor[rgb]{0,0,1}{\textbf{70.7}} \\
				& \cellcolor{mygray}DCP (ours) & \cellcolor{mygray}\textcolor[rgb]{1,0,0}{\textbf{74.9}} & \cellcolor{mygray}\textcolor[rgb]{1,0,0}{\textbf{79.4}} \\ \hline
				\multirow{5}{*}{\multirow{1}{*}[-0.4ex]{ResNet50}} & CANet \cite{shaban2017one} & 66.2 & 69.6 \\												
				& PFENet \cite{tian2020prior} & \textcolor[rgb]{0,0,1}{\textbf{73.3}} & \textcolor[rgb]{0,0,1}{\textbf{73.9}} \\
				& SCL$^\dagger$ \cite{zhang2021self} & 71.9 & 72.8 \\
				& SAGNN \cite{xie2021scale} & 73.2 & 73.3 \\
				& \cellcolor{mygray}DCP (ours) & \cellcolor{mygray}\textcolor[rgb]{1,0,0}{\textbf{75.6}} & \cellcolor{mygray}\textcolor[rgb]{1,0,0}{\textbf{79.7}} \\ \hline
		\end{tabular}}
		\caption{Comparison with state-of-the-arts on PASCAL-$5^i$ in FB-IoU under 1-shot and 5-shot settings. }
		\label{tab:3}
	\end{minipage}
	\vspace{0.1cm}
\end{table*}

\subsection{Training Loss}
To guarantee the richness of the information provided by proxies while preventing extreme situations (\S\textcolor[rgb]{1,0,0}{\,\ref{sec:4.1}}), we introduce additional constraints on the prediction results of self-reasoning schemes for both support and query branches. The total loss for each episode can be written as:
\begin{equation}
\label{eq:8}
\setlength{\abovedisplayskip}{4pt}
{\mathcal{L}_{{\rm{total}}}}{\rm{ = }}{\mathcal{L}_{{\rm{main}}}} + {\lambda _1}{\mathcal{L}_{{\rm{aux1}}}} + {\lambda _2}{\mathcal{L}_{{\rm{aux2}}}},
\setlength{\belowdisplayskip}{4pt}
\end{equation}
where $\mathcal{L}$ represents the binary cross-entropy (BCE) loss function. ${\lambda _1}$ and ${\lambda _2}$, the balancing coefficients, are set to $1.0$ in all experiments. 

\section{Experiments}
\label{sec:5}
\subsection{Setup}
\label{sec:5.1}
\textbf{Datasets.} We evaluate the proposed approach on two standard FSS benchmarks: PASCAL-$5^i$\cite{shaban2017one} and COCO-$20^i$ \cite{nguyen2019feature}. The former is created from PASCAL VOC 2012 \cite{everingham2010pascal} with additional mask annotations from SDS \cite{hariharan2011semantic}, consisting of $20$ semantic categories evenly divided into $4$ folds: $\{{{5^i}}\}_{i = 0}^3$, while the latter, built from MS COCO \cite{lin2014microsoft}, is composed of 80 semantic categories divided into $4$ folds: $\{{{{20}^i}}\}_{i = 0}^3$. Models are trained on $3$ folds and tested on the remaining one in a cross-validation manner. We randomly sample $1,000$ episodes from the unseen fold $i$ for evaluation. 
\vspace{2pt}\\
\textbf{Evaluation Metrics.} Following previous FSS approaches \cite{li2021adaptive,tian2020prior}, we adopt mean intersection-over-union (mIoU) and foreground-background IoU (FB-IoU) as our evaluation metrics, in which the mIoU metric is primarily used since it better reflects the overall performance of different categories.
\vspace{2pt}\\
\textbf{Implementation Details.} Two different backbone networks (\textit{i.e.}, VGG16 \cite{simonyan2014very} and ResNet50 \cite{he2016deep}) are adopted for a fair comparison with existing FSS methods. Following \cite{tian2020prior}, we pre-train these backbone networks on ILSVRC \cite{Russakovsky2015ImageNetLS} and freeze them during training to boost the generalization capability. The setup of pre-processing technology is the same as \cite{tian2020prior}. The SGD optimizer is utilized with a learning rate of $0.005$ for $200$ epochs on PASCAL-$5^i$ and $50$ epochs on COCO-$20^i$. All experiments are performed on NVIDIA RTX 2080Ti GPUs with the PyTorch framework. We report the average results of $5$-runs with different random seeds to reduce the influence of selected support-query image pairs on performance. 

\subsection{Comparison with State-of-the-arts}
\label{sec:5.2}
\begin{figure}[t]
	\vspace{-0.1cm}
	\centering
	\setlength{\abovecaptionskip}{4pt}
	\includegraphics[width=0.85\linewidth]{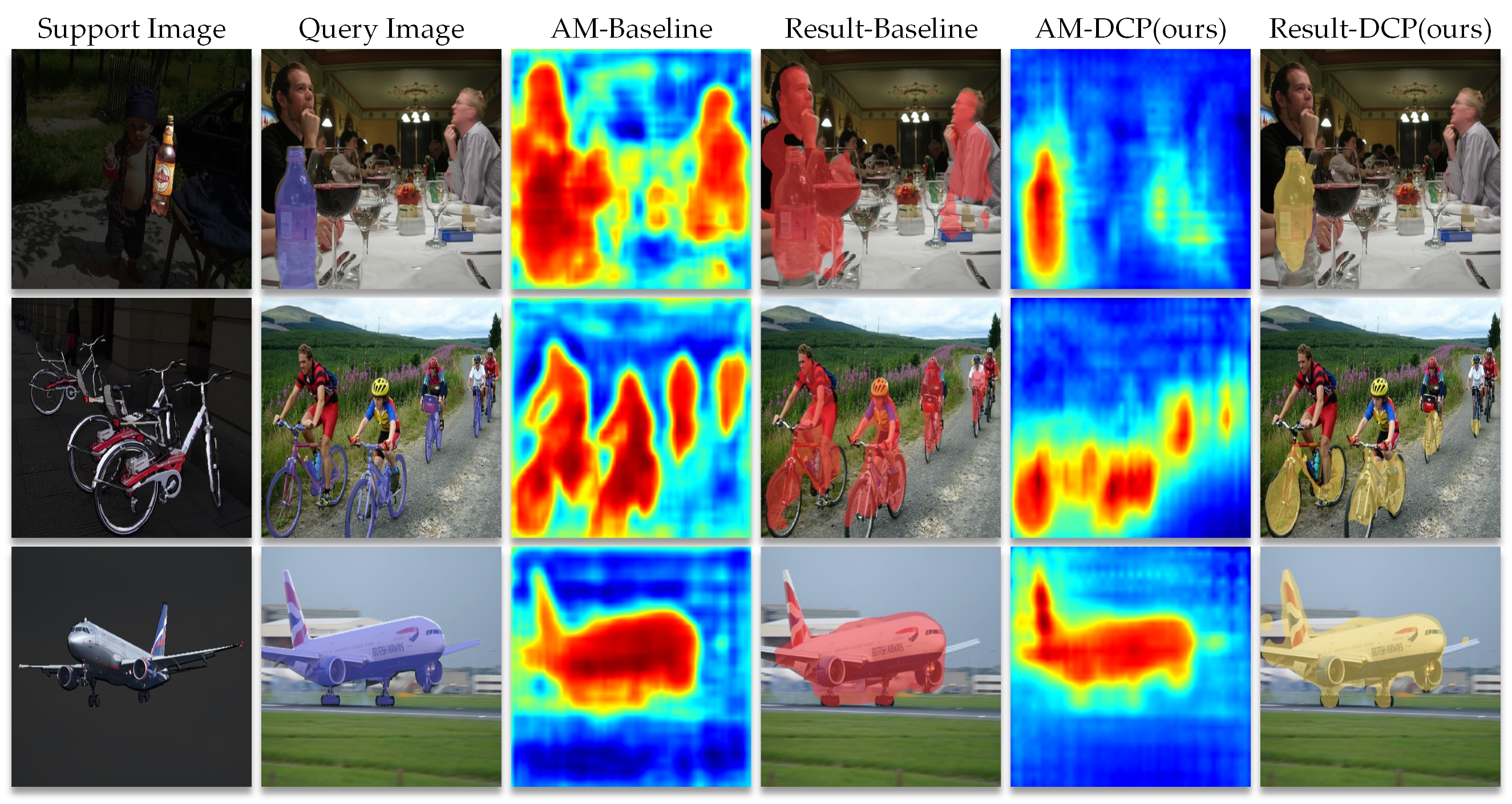}
	\caption{Semantic segmentation results on unseen categories under 1-shot setting. ``AM'' denotes activation map. From left to right, each column represents support images, query images (\textcolor[rgb]{0,0,1}{blue}), baseline AMs, baseline results (\textcolor[rgb]{1,0,0}{red}), our AMs, and our results (\textcolor[rgb]{1,0.78,0.1}{yellow}), respectively. Best viewed in color.}
	\label{fig:5}
	\vspace{0.2cm}
\end{figure}

\begin{figure}[t]
	\centering
	\setlength{\abovecaptionskip}{2pt}
	\includegraphics[width=0.85\linewidth]{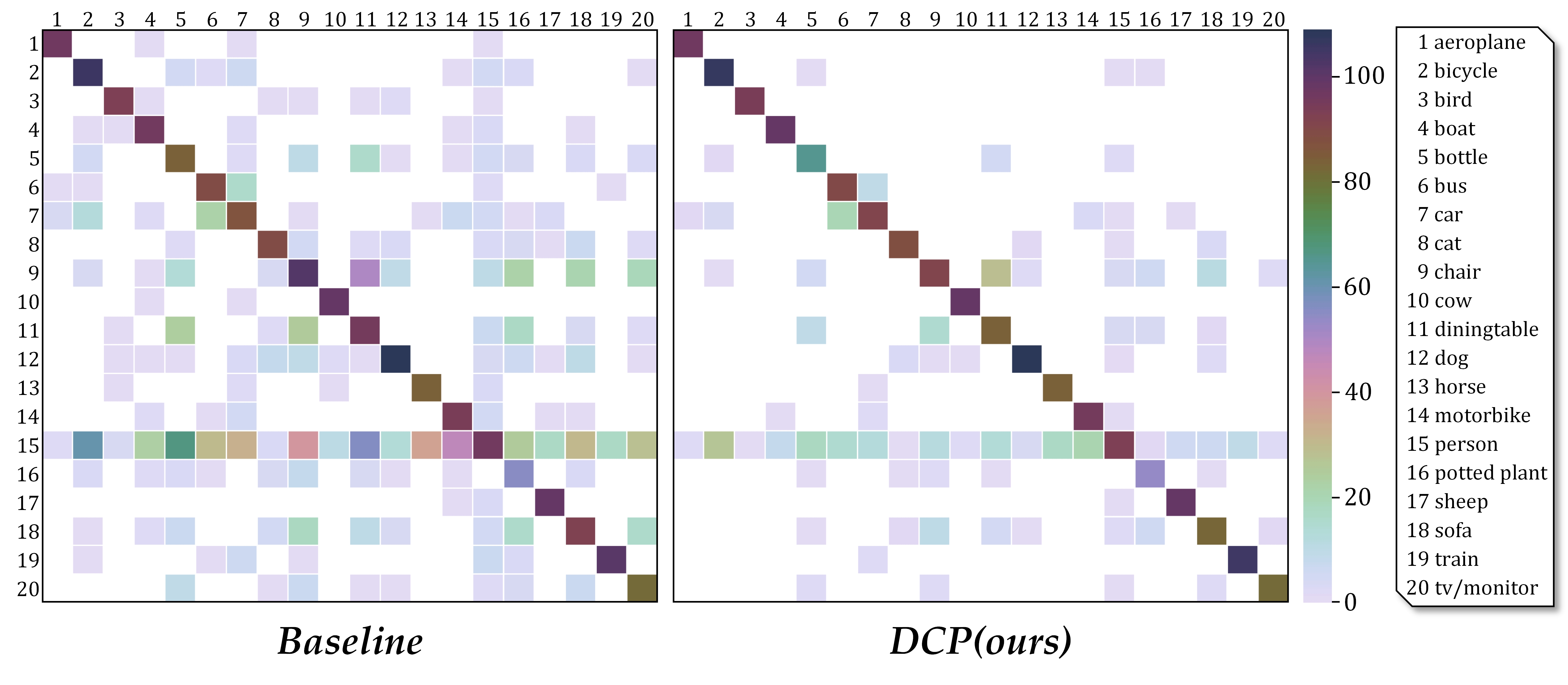}
	\caption{Confusion matrices of baseline approach and our DCP. As can be seen, our DCP effectively reduces semantic confusion between object categories. Best viewed in color.}
	\label{fig:6}
\end{figure}

We evaluate the proposed DCP framework with state-of-the-art methods on two standard FSS datasets. Table\,\textcolor[rgb]{1,0,0}{\ref{tab:1}} presents the 1-shot and 5-shot results assessed under the mIoU metric on PASCAL-$5^i$. It can be observed that our models outperform existing approaches by a sizeable margin, especially under the 5-shot setting, indicating that more appropriate and reliable information is supplied for query image segmentation. Table\,\textcolor[rgb]{1,0,0}{\ref{tab:2}} summarizes the performance of different methods with ResNet50 backbone on COCO-$20^i$, in which the proposed approach sets new state-of-the-arts with a small number of learnable parameters. Specifically, our DCP achieves $2.89\%$ (1-shot) and $3.78\%$ (5-shot) of mIoU improvements over the previous best approach (\textit{i.e.}, PFENet), respectively. The FB-IoU evaluation metric is also included for further comparison, as shown in Table\,\textcolor[rgb]{1,0,0}{\ref{tab:3}}. Once again, the proposed DCP substantially surpasses recent methods. \\
\indent For qualitative analysis, we first visualize the segmentation results of DCP and baseline approach, as illustrated in Figure\,\textcolor[rgb]{1,0,0}{\ref{fig:5}}. It can be found that the false positives caused by irrelevant objects are significantly reduced (see the first two rows). Meanwhile, more accurate segmentation boundaries are obtained, benefiting from the superiority of ${{\bf{p}}_\beta }$ (see the last row). Note that for more details of the baseline approach, please refer to Table\,\textcolor[rgb]{1,0,0}{\ref{tab:4}} in the ablation studies (\S\textcolor[rgb]{1,0,0}{\,\ref{sec:5.3}}). Then we randomly sample $2,000$ episodes from $20$ categories of PASCAL VOC 2012 to draw the confusion matrix, as shown in Figure\,\textcolor[rgb]{1,0,0}{\ref{fig:6}}. The proposed DCP effectively reduces the semantic confusion between object categories and exhibits better segmentation performance. 
\begin{table}[t]
	\centering
	\resizebox{0.8\linewidth}{!}{
		\renewcommand{\arraystretch}{1.1}
		\begin{tabular}{|cc|cccc|c|cc|}
			\hline
			\multicolumn{2}{|c|}{\multirow{1}{*}[-0.2ex]{Prototype}} & \multicolumn{4}{c|}{\multirow{1}{*}[-0.2ex]{Proxy}} & \multirow{2}{*}[-0.2ex]{PDS} & \multirow{2}{*}[-0.2ex]{mIoU} & \multirow{2}{*}[-0.2ex]{FB-IoU} \\ \cline{1-6}
			\multirow{1}{*}[-0.2ex]{$\bf{M}_{\rm{f}}$} & \multirow{1}{*}[-0.2ex]{$\bf{M}_{\rm{b}}$} & \multirow{1}{*}[-0.2ex]{$\bf{M}_{\alpha}$} & \multirow{1}{*}[-0.2ex]{$\bf{M}_{\beta}$} & \multirow{1}{*}[-0.2ex]{$\bf{M}_{\gamma}$} & \multirow{1}{*}[-0.2ex]{$\bf{M}_{\delta}$} &  &  &  \\ \hline \hline
			&  &  &  &  &  &  & \multirow{1}{*}[-0.2ex]{58.52} & \multirow{1}{*}[-0.2ex]{72.46} \\
			\ding{51} &  &  &  &  &  &  & \multirow{1}{*}[-0.2ex]{59.64} & \multirow{1}{*}[-0.2ex]{73.04} \\
			\ding{51} & \ding{51} &  &  &  &  &  & \multirow{1}{*}[-0.2ex]{59.05} & \multirow{1}{*}[-0.2ex]{72.86} \\
			\ding{51} & \ding{51} &  &  &  &  & \ding{51} & \multirow{1}{*}[-0.2ex]{60.12} & \multirow{1}{*}[-0.2ex]{73.20} \\
			&  & \ding{51} & \ding{51} & \ding{51} & \ding{51} & \ding{51} & \multirow{1}{*}[-0.2ex]{60.63} & \multirow{1}{*}[-0.2ex]{73.85} \\
			\ding{51} & \ding{51} & \ding{51} & \ding{51} &  &  & \ding{51} & \multirow{1}{*}[-0.2ex]{60.42} & \multirow{1}{*}[-0.2ex]{73.56} \\
			\ding{51} & \ding{51} &  &  & \ding{51} & \ding{51} & \ding{51} & \multirow{1}{*}[-0.2ex]{60.76} & \multirow{1}{*}[-0.2ex]{74.11} \\
			\ding{51} & \ding{51} & \ding{51} & \ding{51} & \ding{51} & \ding{51} & \ding{51} & \multirow{1}{*}[-0.2ex]{61.31} & \multirow{1}{*}[-0.2ex]{74.87} \\ \hline
	\end{tabular}}
	\caption{Ablation studies on each component.}
	\label{tab:4}
\end{table}
\begin{table}[t]
	\vspace{0.25cm}
	\centering
	\resizebox{0.90\linewidth}{!}{
		\renewcommand{\arraystretch}{1.1}
		\begin{tabular}{|c|ccccc|}
			\hline
			\multirow{1}{*}[-0.2ex]{Fusion Strategy} & \multirow{1}{*}[-0.2ex]{mIoU} & \multirow{1}{*}[-0.2ex]{FB-IoU} & \multirow{1}{*}[-0.2ex]{\#Params.} & \multirow{1}{*}[-0.2ex]{FLOPs} & \multirow{1}{*}[-0.2ex]{Speed} \\ \hline \hline
			\multirow{1}{*}[-0.2ex]{Activation Map} & \multirow{1}{*}[-0.2ex]{61.31} & \multirow{1}{*}[-0.2ex]{74.87} & \multirow{1}{*}[-0.2ex]{0.26M} & \multirow{1}{*}[-0.2ex]{0.24G} & \multirow{1}{*}[-0.2ex]{16.1FPS} \\
			\multirow{1}{*}[-0.2ex]{Tile \& Concat.} & \multirow{1}{*}[-0.2ex]{60.50} & \multirow{1}{*}[-0.2ex]{74.56} & \multirow{1}{*}[-0.2ex]{0.52M} & \multirow{1}{*}[-0.2ex]{0.47G} & \multirow{1}{*}[-0.2ex]{14.9FPS} \\ \hline
	\end{tabular}}
	\caption{Ablation studies on fusion strategies. ``FLOPs'' indicates the computational overhead. ``Speed'' denotes the evaluation of average frame-per-second (FPS) under the 1-shot setting.}
	\label{tab:6}
	\vspace{0.1cm}
\end{table}

\subsection{Ablation Studies}
\label{sec:5.3}
The following ablation studies are performed with VGG16 backbone on PASCAL-$5^i$ unless otherwise stated. \vspace{0pt}\\
\textbf{Proxy \& Prototype.} As discussed in \S\textcolor[rgb]{1,0,0}{\,\ref{sec:4.2}}, the derived proxies, along with foreground/background prototypes, tend to propagate pertinent information for the query branch in the form of activation maps. In Table\,\textcolor[rgb]{1,0,0}{\ref{tab:4}}, we evaluate the impact of each of them on segmentation performance under 1-shot settings. Overall, the following observations could be made: \textbf{(i)} By introducing the foreground activation map ${{\bf{M}}_{\rm{f}}}$, the performance of baseline approach is greatly improved from $58.52\%$ to $59.64\%_{\bf{+1.12}}$, demonstrating the importance of foreground cues for guidance. Furthermore, we argue that such a simple yet powerful variant could serve as a new baseline approach for future FSS works. \textbf{(ii)} The background information provided by mask annotations needs to be used with \textit{caution} since the performance might even deteriorate without PDS (see the third row). \textbf{(iii)} Our proposed proxies show a superior capability compared to the prototypes in facilitating query image segmentation ($60.12\%\,$vs.$\,60.63\%$ mIoU), which is the so-called ``beyond the prototype''. \textbf{(iv)} There is a complementary relationship between the prototype and proxy, the combination of which could further boost performance. \textbf{(v)} Surprisingly, background-related proxies could be better integrated with prototypes (see rows 6-7). One possible reason for this phenomenon is that the interference of irrelevant objects may have a greater influence on the FSS model among many factors, while ${{\bf{M}}_\delta }$ can effectively reduce false positives. \\
\textbf{Parallel Decoder Structure.} After obtaining the activation maps (Eq.\,(\textcolor[rgb]{1,0,0}{\ref{eq:6}})), we first attempt to leverage a simple fusion scheme to guide the query branch, \textit{i.e.}, concatenating ${\bf{F}}_{{\rm{q;f}}}^{{\rm{tmp}}}$ and ${\{ {{{\bf{A}}_l}}\}_{l \in {P_1} \cup {P_2}}}$ and feeding the refined features to a single decoder network. Unfortunately, such a fusion scheme does not offer performance gains, which we attribute to the \textit{information confusion} between the expanded foreground prototype ${\zeta _{H \times W}}\left( {{{\bf{p}}_{\rm{f}}}} \right)$ and background-related activation maps (${{\bf{M}}_\gamma}$ and ${{\bf{M}}_\delta}$). To this end, we devise a unique parallel decoder structure (PDS) where proxies/prototypes with similar attributes are integrated, yielding a superior result compared to the previous scheme (see rows 3-4 of Table\,\textcolor[rgb]{1,0,0}{\ref{tab:4}}). \vspace{0pt}\\
\textbf{Efficiency.} We also evaluate the efficiency of different fusion strategies between derived proxies ${\{{{{\bf{p}}_m}}\}_{m \in {P_1}}}$ and query features ${{\bf{F}}_{\rm{q}}}$, as presented in Table\,\textcolor[rgb]{1,0,0}{\ref{tab:6}}. Please note that the first row represents the previously discussed scheme (see Eq.\,(\textcolor[rgb]{1,0,0}{\ref{eq:7}})), while the second row denotes the scheme that concatenates query features with the expanded proxies, similar to Eq.\,(\textcolor[rgb]{1,0,0}{\ref{eq:6}}). The experiment results indicate that the first fusion strategy exhibits remarkable segmentation performance with fewer parameters ($0.26$M) and faster speed ($16+\,$FPS), demonstrating the effectiveness of activation maps for guidance. 

\vspace{-0.3cm}
\section{Conclusion}
\vspace{-0.1cm}
In this paper, we have proposed an efficient framework with the spirit of divide-and-conquer, which performs a self-reasoning scheme to derive support-induced proxies for tackling the typical challenges of FSS tasks. Moreover, a novel parallel decoder structure was introduced to further boost the discrimination power. Qualitative and quantitative results show that the proposed proxies could provide appropriate and reliable information to facilitate query image segmentation, surpassing conventional prototype-based approaches by a considerable margin, \textit{i.e.}, the so-called ``beyond prototype''.

\vspace{-0.3cm}
\section*{Acknowledgements}
\vspace{-0.1cm}
{This work was supported in part by the National Natural Science Foundation of China under Grants 62136007 and U20B2065, in part by the Shaanxi Science Foundation for Distinguished Young Scholars under Grant 2021JC-16, and in part by the Fundamental Research Funds for the Central Universities.}

\clearpage
{\small
	\bibliographystyle{named}
	\bibliography{DCP}
}

\end{document}